\documentclass{article}
\usepackage{spconf,amsmath,graphicx}

\title{A Method to Suppress Facial Expressions in Posed and Spontaneous Videos}
\name{Ghada Zamzmi, Gabriel Ruiz, Matthew Shreve, Dmitry Goldgof, Rangachar Kasturi, and Sudeep Sarkar}
\address{Department of Computer Science and Engineering \\
University of South Florida \\
Tampa, Florida, USA}
\usepackage{float}
\begin{document}
%
\maketitle
\begin{abstract}
We address the problem of suppressing facial expressions in videos because expressions can hinder the retrieval of important information in applications such as face recognition. To achieve this, we present an optical strain suppression method that removes any facial expression without requiring training for a specific expression. For each frame in a video, an optical strain map that provides the strain magnitude value at each pixel is generated; this strain map is then utilized to neutralize the expression by replacing pixels of high strain values with pixels from a reference face frame. Experimental results of testing the method on various expressions namely happiness, sadness, and anger for two publicly available data sets (i.e., BU-4DFE and AM-FED) show the ability of our method in suppressing facial expressions. 
\end{abstract}
\section{Introduction}
\label{sec:intro}
Facial expressions manipulation has been an active area of research in recent years \cite{brizzi2014optical,gass2011warp,theobald2009mapping}. Several directions such as expression replacing, transferring, exaggerating, and suppressing have been investigated. The presence of facial expressions can provide a rich source of information for human behavior analysis and communication. However, facial expressions can also hinder information retrieval in some cases and affect the performance of approaches that use face as a biometric. Therefore, developing suppression methods that generate a neutral expression can be helpful. These methods can be utilized in various scenarios such as a pre-processing step to face recognition, continuous authentication, and human-computer interaction. 

Expression manipulation and neutralization to improve face recognition has gained more attention recently. The existing approaches in this area can be divided into: geometric based approaches \cite{zhang2006geometry,tan2009face}, model/template based approaches \cite{amberg2008expression,chu20143d,theobald2009mapping}, tensor based approaches \cite{yang2012facial,yang2011expression}, and wrap based approaches \cite{gass2011warp}. The main limitation of these approaches is that the facial annotations and the region segmentation are either manual/semi-automated or require pre-training for model fitting.  

This paper extends our previous work \cite{brizzi2014optical}, which proposed a method to suppress the smile expression for 4 subjects, to include other expressions and larger datasets (i.e., BU-4DFE \cite{yin20063d} and AM-FED \cite{mcduff2013affectiva}). The main advantages of the presented algorithm are as follows: 1) the algorithm has the ability to suppress any expression without requiring prior knowledge of the data or training for a specific expression, and 2) it is simple and has low cost since it applies the suppression directly to the frames that contain the target face rather than transferring the facial information into a separate model. The experimental results of the proposed method on various expressions namely happiness, sadness, and anger show its ability to suppress and reduce facial expressions.

\section{Proposed Method}
\label{sec:method}
In this section, we provide a presentation of our suppression algorithm, which consists of three main stages: 1) facial landmarks detection and tracking, 2) expression suppression based on strain analysis, 3) post-processing stage to smooth the suppressed expression. Each of these stages is described below. 

A subspace constrained mean shift face tracker \cite{saragih2009face} that automatically locates and tracks 66 facial points is applied for facial landmarks detection and tracking. The extracted points are used then to align the face and crop it.

In the second stage, we computed the optical flow vectors between consecutive video frames and used these vectors to derive the optical strain as described in \cite{brizzi2014optical}. The optical strain can be defined as the measure of an object's deformations. In case of facial expressions, it measures the facial tissues' deformations, which are caused by an expression. The strain magnitude, derived from optical flow vectors and computed as described in \cite{brizzi2014optical}, is then normalized to [0,255] and used to generate a strain map. Figure 1 provides an illustration of the normalized strain values (strain map) of a target face across an expression event.

After the strain map is generated, it is used to identify the high strain pixels that need to be replaced based on a simple thresholding. The result is a binary mask that indicates which pixels should be replaced from the reference frame (i.e., frame of neutral expression or lowest strain values) to the current frame. We set the value of this threshold to any number that exceeds the lower 10 percent of values in the strain map. 

\begin{figure}[t]
\begin{center}
\includegraphics[width=2.5cm, height=2.5cm]{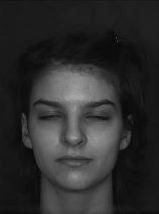}
\includegraphics[width=2.5cm, height=2.5cm]{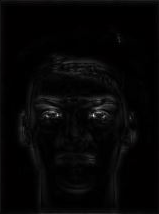}

\includegraphics[width=2.5cm, height=2.5cm]{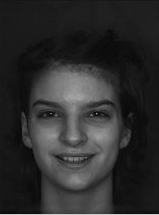}
\includegraphics[width=2.5cm, height=2.5cm]{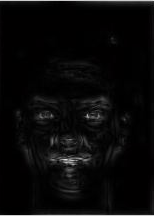}
\end{center}

\begin{center}
\caption{The strain map across different frames (10, 1st row and 52, 2nd row) of a happiness expression.}
\end{center}
\end{figure}

Finally, the post-processing stage consists of two level of smoothing to reduce the artifacts generated as a result of suppressing the expression. The first level of smoothing is applied to the edges of the masked areas to reduce the artifacts of the pixels' masking process using a 2D median smoothing algorithm. The second level of smoothing is used to minimize the false edges on the skin by performing a standard smoothing algorithm on the entire face.

\section{Study Design}
\label{sec:study}
\subsection{Dataset}
Video sequences of two publicly-available datasets were used to evaluate our strain-based method for expression suppression. The first dataset is SUNY Binghampton BU-4DFE \cite{yin20063d} dataset that contains 101 subjects performing 6 different posed prototypical expressions. Each of those subjects has a total of 6 video sequences, which gives a total of 606 video sequences. The second dataset is AM-FED \cite{mcduff2013affectiva} in the wild dataset that contains 242 facial videos collected in real world conditions. The videos have spontaneous facial expressions of subjects recorded while watching amusing Super Bowl commercials. 

\subsection{Study procedure}
In order to evaluate our suppression algorithm, two expression detectors were applied before and after suppressing the expression. The first detector is the OpenCV Haar-cascade object detector, which has been trained to detect smiles. This detector returns the smile's intensity in each frame of a video sequence as a normalized value between 0 and 99. The second expression detector is the FACET Module from the iMotions Attention Tool version 5.3 \cite{iMotions}. This commercial detector was used to detect sadness and anger expressions. Given a subject's face, the detector reports an evidence value that indicates how likely the subject is expressing any of the expressions at a given frame.

\section{Experiments and Results}
\label{sec:experiments}
We report the results of applying the strain-based suppression algorithm on BU-4DFE posed facial expression dataset and AM-FED spontaneous facial expression dataset. We tested the algorithm on BU-4DFE with three expressions (i.e., happiness, sadness, and anger) and AM-FED dataset with only happiness expression since subjects of this dataset only show the happiness expression. The subsections below present the experimental results of applying the strain-based suppression algorithm and its evaluation using expression detectors.  

\begin{figure}[t]
\begin{center}
\includegraphics[width=2.5cm, height=2.5cm]{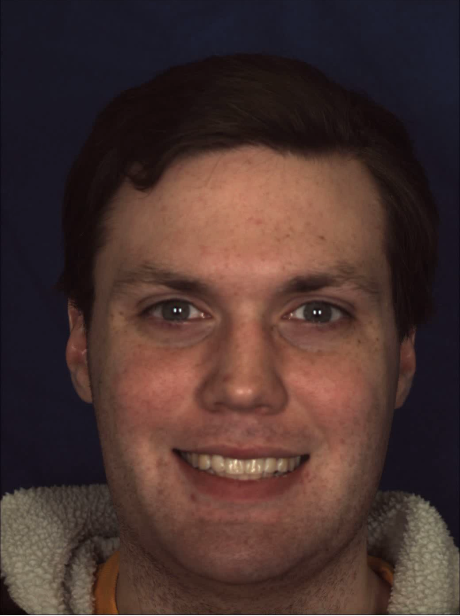}
\includegraphics[width=2.5cm, height=2.5cm]{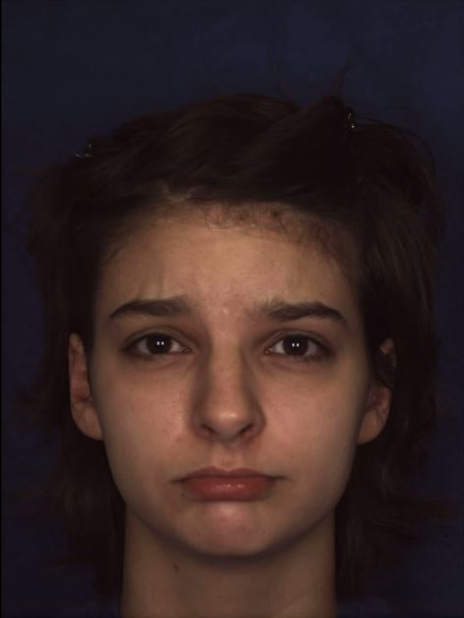}

\includegraphics[width=2.5cm, height=2.5cm]{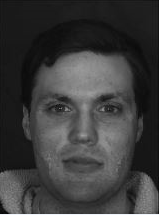}
\includegraphics[width=2.5cm, height=2.5cm]{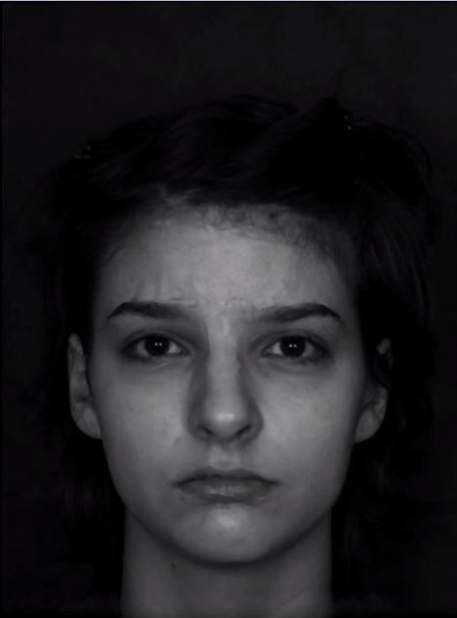}

\begin{center}
\caption{Expression Suppression on BU-4DFE dataset}
\end{center}

\includegraphics[width=2.5cm, height=2cm]{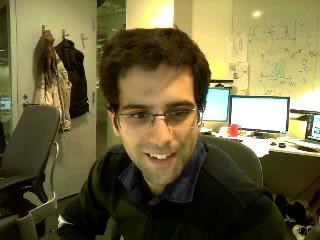}
\includegraphics[width=2.5cm, height=2cm]{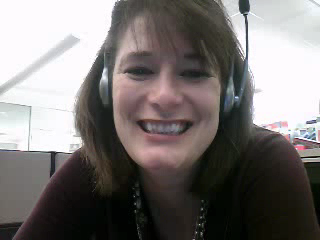}
\includegraphics[width=2.5cm, height=2cm]{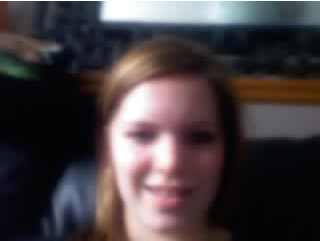}

\includegraphics[width=2.5cm, height=2cm]{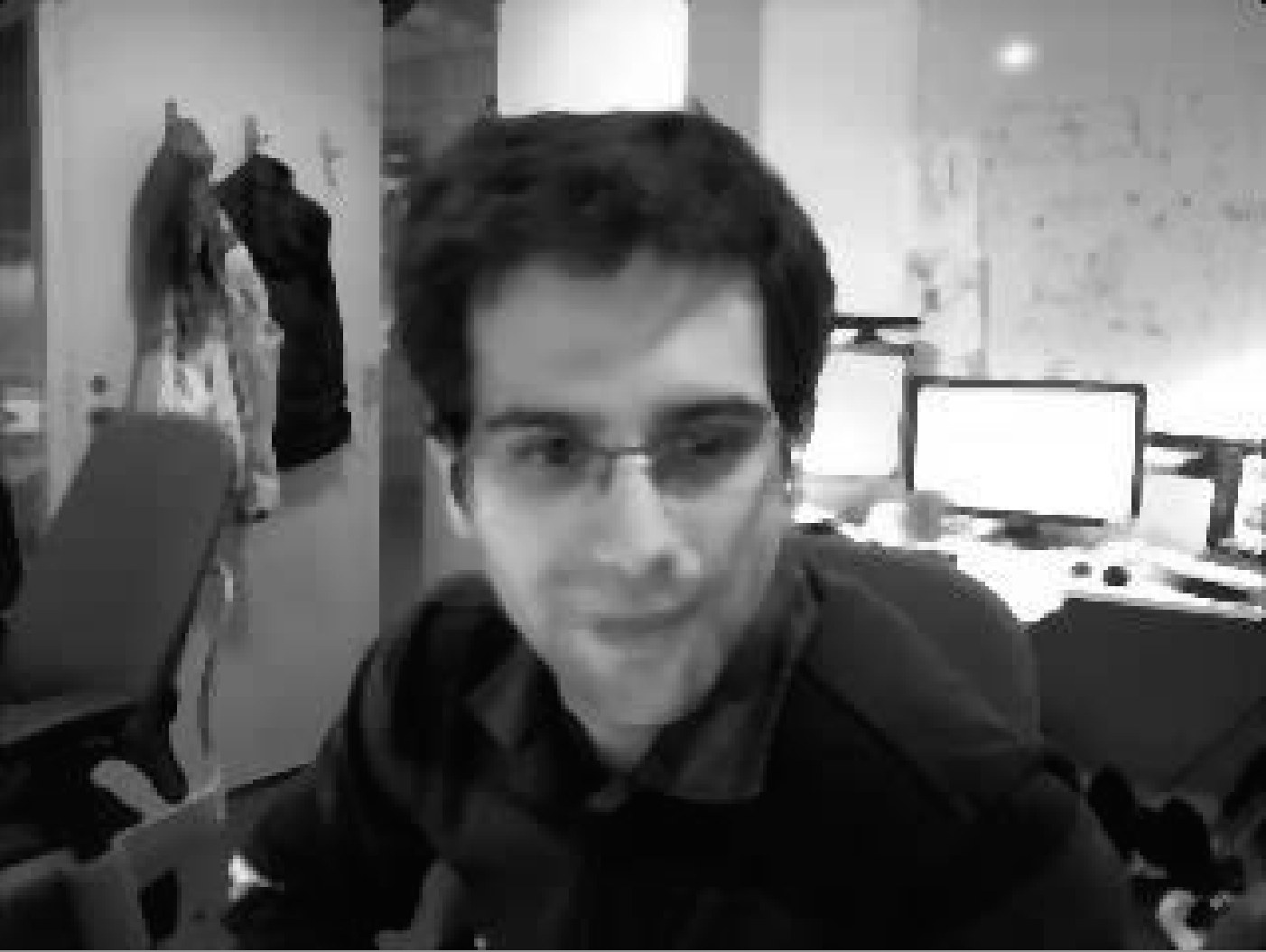}
\includegraphics[width=2.5cm, height=2cm]{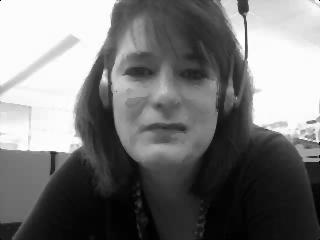}
\includegraphics[width=2.5cm, height=2cm]{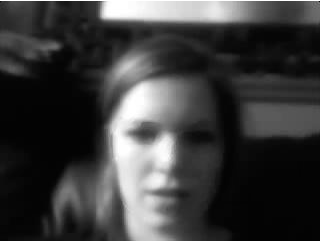}

\begin{center}
\caption{Expression Suppression on AM-FED dataset} 
\end{center}
\end{center}
\end{figure}

Figure 2 depicts the suppression of happiness and sadness for two subjects in BU-4DFE dataset. Figure 3 shows the results of suppressing the spontaneous happiness expression for three subjects in AM-FED dataset. For each subject, the first row represents a frame from the original video and the second row represents the same frame from the generated suppressed video. The third column in Figure 3 shows the ability of our algorithm to suppress the spontaneous happiness expression in low resolution videos.

\subsection{Algorithm Evaluation Using Expression Detectors}
To evaluate the effectiveness of our algorithm in suppressing the expression, OpenCV happiness detector and FACET expressions' detector were utilized to measure the expression's intensity before and after applying the algorithm. OpenCV detector was used to measure the happiness intensity for both BU-4DFE and AM-FED datasets while FACET detector was used to measure the intensity of other expressions in BU-4DFE dataset. 

Before we proceed further in presenting the results, it is important to mention that applying the strain-based suppression algorithm on video sequences gives us one of the three following cases: 1) the algorithm completely suppresses the expression, 2) it reduces the expression intensity, and 3) it increases the expression intensity. The results of evaluating the suppression algorithm for happiness expression in AM-FED dataset and for happiness, sadness, and anger expressions in BU-4DFE dataset are summarized below.

\subsubsection{Expression: Happiness}
Tables 1 and 2 summarize the results of evaluating the algorithm on a set of video sequences depicting the happiness expression. 

Table 1 shows the three cases of applying OpenCV detector to AM-FED dataset.  The table shows that happiness expression for 21\% of all videos in AM-FED dataset was completely suppressed and it was reduced by 50\% for 64\% videos. However, the algorithm failed to reduce the expression but instead it increased its intensity in around 15\% of videos. We think this can be attributed to the facial distortions (i.e., artifacts) caused by the pixels replacement. We believe this issue can be solved by using better facial alignment methods as well as advanced smoothing techniques. Figure 4 illustrates the performance of OpenCV detector before and after suppression for three subjects in AM-FED dataset using a Receiver Operating Characteristic (ROC) curve. This curve shows a noticeable reduction in the performance of OpenCV detector after applying the suppression algorithm.

\begin{table}[H]
\centering
\caption{OpenCV Results for AM-FED Dataset}
\label{my-label}
\begin{tabular}{| c | c | c | }
\hline
 Case & \% of videos & \% changes  \\ 
\hline 
Smile Removed  & \centering 21\% & 100\% \\
Smile Reduced & \centering 64\% & 50\% \\
Smile Increased & \centering 15\% & 10\% \\
\hline
\end{tabular}
\end{table}

Table 2 shows the three cases of applying OpenCV detector to BU-4DFE dataset. Figure 5 shows intensity of the happiness expression in an original video sequence (green) of BU-4DFE dataset, which is completely suppressed after applying the suppression algorithm as illustrated by the blue line. 

\begin{figure}[t]

\begin{center}
\includegraphics [width=6cm, height=4cm]{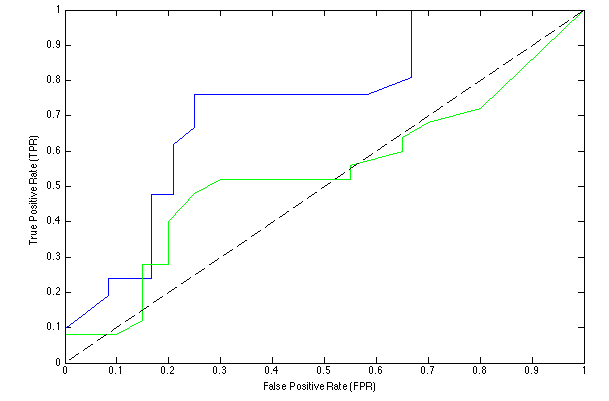}
\caption{ROC curve for OpenCV detector before (blue) and after (green) suppression. } 
\end{center}
\end{figure}

\begin{figure}[t]
\begin{center}
\centering
\includegraphics[width=6.5cm, height=4.5cm]{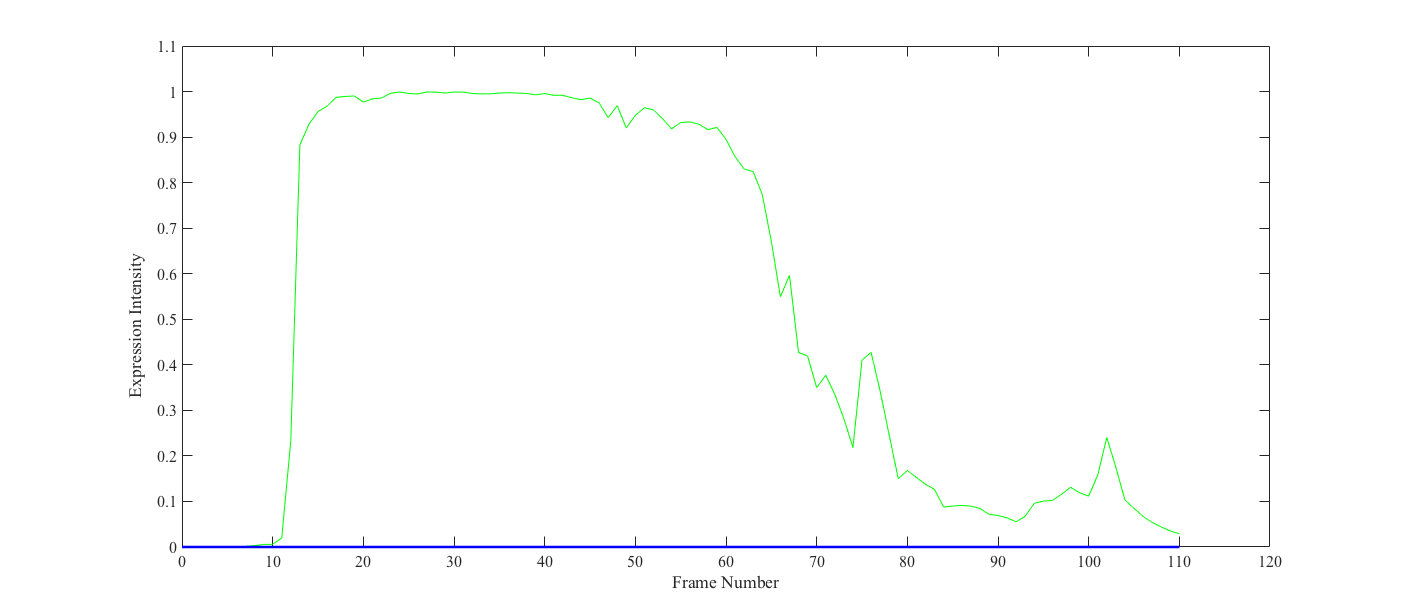}
\centering
\caption{Intensity of happiness before (green) and after (blue) the suppression; the expression starts at frame 13 and ends at 75.}
\end{center}
\end{figure}

\begin{table}[H]
\centering
\caption{OpenCV Results for BU-4DFE Dataset}
\label{my-label}
\begin{tabular}{| c | c | c | }
\hline
 Case & \% of videos & \% changes  \\ 
\hline 
Smile Removed  & \centering 59\% & 100\% \\
Smile Reduced & \centering 38\% & 75\% \\
Smile Increased & \centering 3\% & 5\%\\
\hline
\end{tabular}
\end{table}

\subsubsection{Expression: Sadness}
As shown in Table 3, the sadness expression is reduced in approximately 87\% of the subjects and increased in the remaining. Figure 6 displays the intensity of the sadness expression in an original video sequence (green curve) of BU-4DFE Dataset, which is reduced after applying the algorithm as illustrated by the blue curve. The figure shows that the expression’s intensity is reduced to below 50\% in most frames.

\begin{table}[t]
\centering
\caption{Sadness Results Using FACET with BU-4DFE }
\label{my-label}
\begin{tabular}{| c | c | c | }
\hline
 Case & \% of videos & \% changes  \\ 
\hline 
Sadness Removed  & \centering 0\% & 100\% \\
Sadness Reduced & \centering 87\% & 60\% \\
Sadness Increased & \centering 13\% & 22\% \\
\hline
\end{tabular}
\end{table}

\subsubsection{Expression: Anger}
The suppression algorithm was able to reduce the anger expression by 40\% for most of the subjects and completely suppressed the expression in about 6\% of the subjects. The results are summarized in Table 4. Figure 7 displays the intensity of the anger expression in an original video sequence (green curve) as compared to the intensity in the suppressed video sequence (blue curve).

\begin{table}[t]
\centering
\caption{Anger Results Using FACET with BU-4DFE}
\label{my-label}
\begin{tabular}{| c | c | c | }
\hline
 Case & \% of videos & \% changes  \\ 
\hline 
Anger Removed  & \centering 6\% & 100\% \\
Anger Reduced & \centering 88\%  & 40\% \\
Anger Increased & \centering 6\%  & 36\% \\
\hline
\end{tabular}
\end{table}

\begin{figure}[t]
\begin{center}
\centering
\includegraphics[width=6.5cm, height=4.5cm]{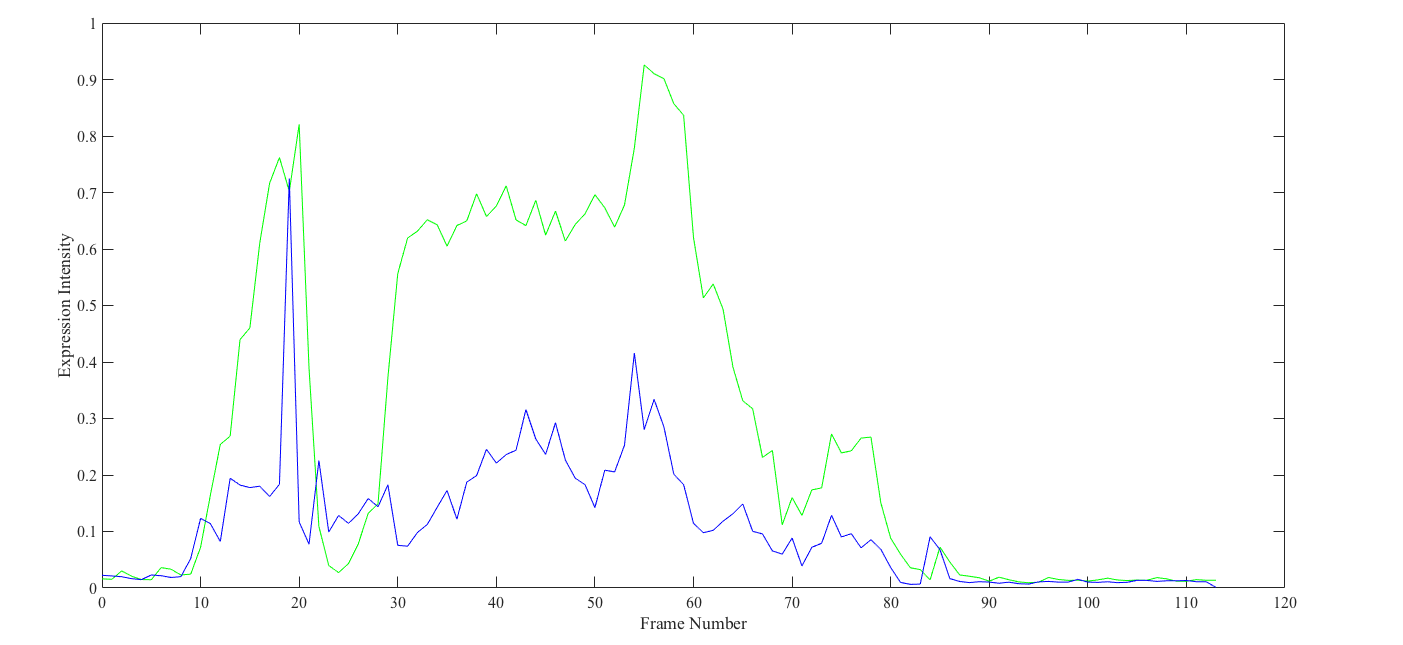}

\caption{ Intensity of sadness before (green) and after (blue) the algorithm; the expression starts at frame 12 and ends at 76.}
\end{center}
\end{figure}

\begin{figure}[t]
\begin{center}
\centering
\includegraphics[width=6.5cm, height=4.5cm]{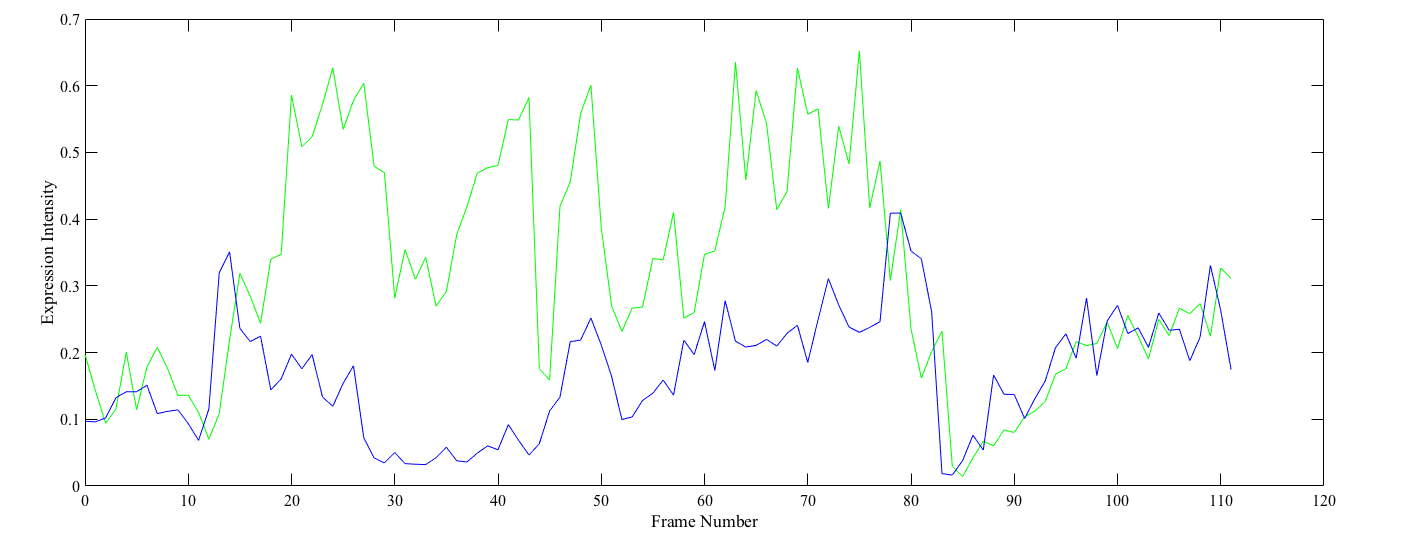}
\centering
\caption{Intensity of anger before (green) and after (blue) the algorithm; the expression starts at frame 11 and ends at 76.}
\end{center}
\end{figure}

\section{Conclusions}
\label{sec:majhead}
This paper presents a strain-based method for suppressing facial expressions in videos. Evaluating the algorithm on two publicly-available datasets (AM-FED spontaneous expression dataset and BU-4DFE posed dataset) proves the capability of the presented algorithm in reducing if not completely suppressing facial expressions. The algorithm yields 75\% reduction in the happiness expression in BU-4DFE dataset. The reduction of the same expression in AM-FED (the wild dataset) was about 50\%. The lower reduction rate in case of spontaneous facial expressions can be attributed to the nature of these expressions (i.e., spontaneous facial expressions tend to be less intense and dynamic comparing to the posed expressions). 

The algorithm was able to reduce anger and sadness expressions significantly although the average reductions are lower than that of the happiness expression. It is possible to attribute the lesser effectiveness of suppression for anger and sadness to the subjects' representation of those expressions. Smiling was the universal representation for happiness, while the expressions for anger and sadness were more variable among the subjects. This does not affect the suppression algorithm, but may cause expression detectors to yield different results. The results of this work are encouraging; they prove the effectiveness of using facial strain for expression suppression and open new directions for future works. 

In future, we plan to follow three main directions. First, we will investigate the most recent facial tracker to improve the facial alignment and registration. We believe improving the facial alignment would improve the performance since our algorithm depends on pixel replacement; a high accuracy facial alignment can decrease the artifacts generated by pixel replacement. Second, we will evaluate our algorithm on other expressions such as surprise and disgust using expression detectors designed for detecting these emotions. Finally, we plan to compare our algorithm with the state of the arts in expression suppression.


\end{document}